%% file: main.tex
\documentclass{article} 
\usepackage{iclr2026_conference,times}

\input{math_commands.tex}

\usepackage{hyperref}
\usepackage{url}
\usepackage{graphicx}
\usepackage{caption}
\usepackage{subcaption}
\usepackage{psfrag}
\usepackage{array}
\usepackage{multirow}
\usepackage{algorithm}
\usepackage{algorithmicx}
\usepackage{algpseudocode}
\usepackage{booktabs}
\usepackage{comment}
\usepackage{amsmath}

\usepackage{xcolor}
\usepackage{amssymb}
\usepackage{amsfonts}
\usepackage{amssymb}
\usepackage{setspace}
\usepackage{textcomp}

\usepackage{cleveref}
\usepackage{titlesec}

\titlespacing*{\section}{0pt}{1.2ex plus 0.4ex minus 0.2ex}{0.6ex}
\titlespacing*{\subsection}{0pt}{0.9ex plus 0.3ex minus 0.2ex}{0.4ex}
\titlespacing*{\subsubsection}{0pt}{0.4ex plus 0.2ex minus 0.1ex}{0.2ex}
\title{Function-Space Decoupled Diffusion for \\ 
Forward and Inverse Modeling in Carbon \\
Capture and Storage}

\author{
Xin Ju$^{1}$ \;
Jiachen Yao$^{2}$ \;
Anima Anandkumar$^{2}$ \;
Sally M. Benson$^{1}$ \;
Gege Wen$^{3}$
\\[0.6em]
$^{1}$Stanford University \;
$^{2}$California Institute of Technology \\
$^{3}$Imperial College London
}

\iclrfinalcopy 
\begin{document}

\maketitle

\begin{abstract}
Accurate characterization of subsurface flow is critical for Carbon Capture and Storage (CCS) but remains challenged by the ill-posed nature of inverse problems with sparse observations.
We present Function-space \emph{Decoupled} Diffusion Posterior Sampling (Fun-DDPS), a generative framework that combines function-space diffusion models with differentiable neural operator surrogates for both forward and inverse modeling.
Our approach learns a prior distribution over geological parameters (geomodel) using a single-channel diffusion model, then leverages a Local Neural Operator (LNO) surrogate to provide physics-consistent guidance for cross-field conditioning on the dynamics field.
This decoupling allows the diffusion prior to robustly recover missing information in parameter space, while the surrogate provides efficient gradient-based guidance for data assimilation.
We demonstrate Fun-DDPS on synthetic CCS modeling datasets, achieving two key results:
(1) For forward modeling with only 25\% observations, Fun-DDPS achieves 7.7\% relative error compared to 86.9\% for standard surrogates (an 11$\times$ improvement), proving its capability to handle extreme data sparsity where deterministic methods fail.
(2) We provide the first rigorous validation of diffusion-based inverse solvers against asymptotically exact Rejection Sampling (RS) posteriors. Both Fun-DDPS and the joint-state baseline (Fun-DPS) achieve Jensen-Shannon divergence $<0.06$ against the ground truth. 
Notably, Fun-DDPS produces physically consistent realizations free from the high-frequency artifacts observed in joint-state baselines, achieving this with 4$\times$ improved sample efficiency compared to rejection sampling.
\end{abstract}

\section{Introduction}
\label{sec:intro}

Carbon capture and storage (CCS) is a critical technology for mitigating anthropogenic climate change~\citep{pacala2004stabilization, international2020energy}.
The safety and efficacy of gigaton-scale storage depend on two complementary computational tasks: \textit{forward modeling}, to forecast CO$_2$ plume migration and pressure buildup, and \textit{inverse modeling}, to characterize subsurface geological heterogeneity from sparse monitoring data.
However, robust uncertainty quantification in this domain faces a fundamental bottleneck: subsurface parameters are high-dimensional and non-Gaussian, while the governing multiphase flow equations are computationally expensive to solve.

Current data assimilation (DA) methods struggle to effectively tackle these challenges.
Traditional ensemble-based methods, such as the Ensemble Kalman Filter (EnKF) and Ensemble Smoother (ES-MDA), remain the industry standard but rely on Gaussian assumptions that fail to capture complex geological features such as discrete facies or channelized reservoirs~\cite{Nejadi2012, Emerick2013b}.
Conversely, rigorous Bayesian sampling methods including Markov Chain Monte Carlo (MCMC)~\cite{evensen2003ensemble, evensen2004sampling, vrugt2013hydrologic} avoid these assumptions but are prohibitively expensive, often requiring thousands of high-fidelity simulations that are infeasible for large-scale 3D models~\citep{Oliver1997}.
While deep learning surrogates, such as Fourier Neural Operators (FNO), have successfully accelerated forward simulations by orders of magnitude~\citep{li2020fourier, wen2022u}, they are typically deterministic and do not inherently address the ill-posed inverse problem directly.

To bridge this gap, generative diffusion models have emerged as powerful priors for scientific inverse problems~\citep{song2020score, chung2022diffusion, yang2023diffusion, yao2025guided}.
However, applying them to CCS encounters a critical limitation in existing \textit{joint-state} architectures, which learn the joint distribution of geological parameters and dynamic states $p(\boldsymbol{m}, \boldsymbol{s})$ together.
We observe that joint training often leads to physical inconsistency, as the model learns statistical correlations rather than explicit physical laws, especially when paired training data is limited~\citep{lin2025ddis}.

In this work, we propose \textbf{Fun-DDPS} (Function-space \emph{Decoupled} Diffusion Posterior Sampling), a novel framework that decouples the learning of geological priors from the approximation of flow physics.
Our approach learns a prior distribution over geological parameters $p(\boldsymbol{m})$ using a single-channel function-space diffusion model, and independently trains a neural operator surrogate $\mathcal{L}_\phi$ to approximate the forward physics.
During inference, we employ the differentiable neural surrogate to backpropagate gradients from sparse observations directly into the diffusion generation process.
This decoupled architecture offers three key contributions:
\begin{enumerate}
    \item \textbf{Robust Forward Modeling:} By leveraging the generative prior to reconstruct missing data, Fun-DDPS achieves 7.7\% relative error on forward tasks with 25\% data coverage, compared to 86.9\% error for standard surrogates.
    \item \textbf{Physics-Consistent Inversion:} We demonstrate that guiding the diffusion process via a physics-based surrogate eliminates the high-frequency artifacts common in joint-state models, producing geologically realistic posterior samples even with limited observations.
    \item \textbf{Rigorous Validation:} We provide the first rigorous benchmark of diffusion-based inversion against asymptotically exact Rejection Sampling (RS) posteriors, achieving high statistical accuracy (JS divergence $<0.06$) with a $4\times$ reduction in computational cost.
\end{enumerate}

\begin{figure}[t]
    \centering
    \includegraphics[width=\columnwidth]{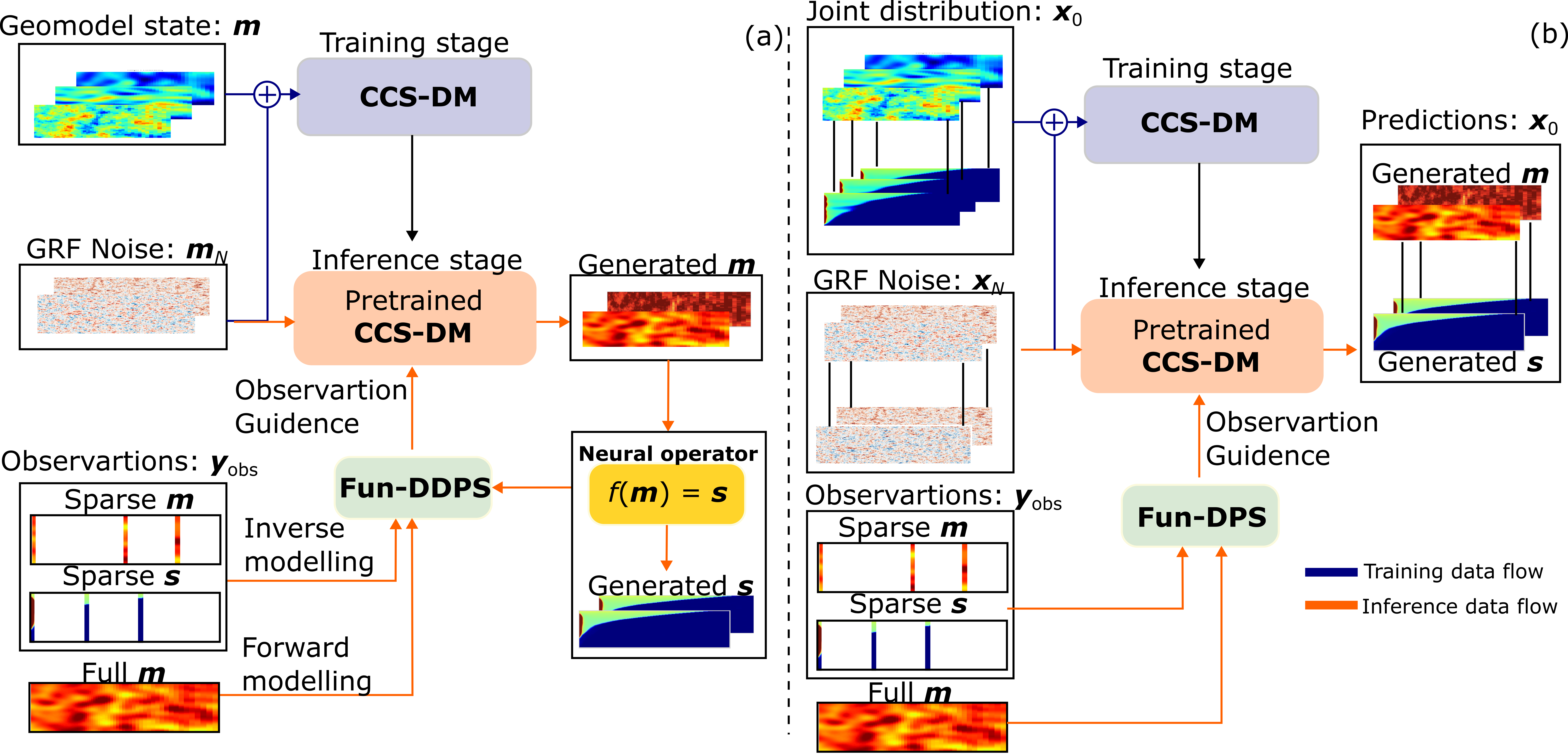}
    \caption{
        Comparison of decoupled (Fun-DDPS) vs.\ joint-state (Fun-DPS) architectures.
        \textbf{(a) Fun-DDPS (Decoupled):} Training uses only geomodel samples $\boldsymbol{m}$ to learn the prior $p(\boldsymbol{m})$; a separate neural operator surrogate $\mathcal{L}_\phi$ learns the forward mapping. During inference, the geomodel diffusion model generates $\boldsymbol{m}$, and $\mathcal{L}_\phi$ maps it to dynamics $\boldsymbol{s}$. For inverse problems, dynamics observations guide the diffusion \textit{through} the surrogate gradient.
        \textbf{(b) Fun-DPS (Joint-state):} Training requires paired data $(\boldsymbol{m}, \boldsymbol{s})$ to learn the joint distribution $p(\boldsymbol{m}, \boldsymbol{s})$. During inference, the joint model generates both fields simultaneously, with observations guiding the corresponding channels directly.}
    \label{fig:framework_schematic}
\end{figure}

\section{Related Work}
\label{sec:related_work}

\textbf{Inverse Problems in Subsurface Systems.}
The inverse problem in hydrology and reservoir engineering is classically addressed via variational or ensemble methods.
Gradient-based variational approaches~\citep{lewis2006dynamic} are efficient but often require adjoint code, which is unavailable for many black-box simulators.
Derivative-free methods such as ES-MDA~\citep{Emerick2013b} are widely used but suffer from the ``Gaussian limitation," often smoothing out sharp geological interfaces critical for flow connectivity.
While recent work has integrated deep generative models (GANs, VAEs) into ensemble smoothers to parameterize non-Gaussian priors~\citep{misra2023massive, forghani2022variational, misra2024generative, teng2025generating}, these approaches typically rely on low-dimensional latent spaces that may limit the expressivity of the recovered fields.
%

\textbf{Deep-Learning-Based Surrogates for Inverse Problems.} 
Recent advances have introduced various deep learning-based (DL) strategies to address the data assimilation challenges. 
One popular approach involves using DL models as fast surrogates to replace conventional reservoir simulators for forward simulation.
These surrogates, including Recurrent Residual U-Nets, Fourier neural operators (FNO), and graph neural networks (GNN), effectively reduce forward simulation costs~\cite{tang2020deep, tang2022deep, wen2022accelerating, wen2022u, ju2024learning}.
This speedup makes many computationally intensive DA methods feasible.
DL surrogates can facilitate Bayesian methods that treat geological hyperparameters (e.g., mean permeability, correlation lengths) as uncertain variables for CCS applications. 
However, even with fast surrogates, data assimilation remains within conventional frameworks where ESMDA is used to calibrate reservoir parameters, meaning the reliance on Gaussian assumptions in the inversion step is not fully resolved.

\textbf{Diffusion Models for Inverse Problems.}
Denoising Diffusion Probabilistic Models (DDPMs)~\citep{ho2020denoising} currently define the state-of-the-art in generative modeling.
For inverse problems, methods such as \textit{Diffusion Posterior Sampling} (DPS)~\citep{chung2022diffusion, huang2024diffusionpde} allow for conditional generation by using the gradient of a forward operator to guide the reverse diffusion process.
In the scientific domain, this has been extended to infinite-dimensional function spaces~\citep{lim2023score, kovachki2021neural}.
Most relevant to our work is the \textit{Decoupled Diffusion Inverse Solver} (DDIS)~\citep{lin2025ddis}, which showed the failure modes of joint-state training.
We extend this decoupled idea to function-space CCS modeling, specifically addressing the extreme sparsity of dynamic monitoring data ($<1\%$ coverage) and providing a rigorous quantitative benchmark against the ground truth posterior obtained via Rejection Sampling.

\section{Methodology}
\label{sec:methodology}

We propose \textbf{Fun-DDPS} (schematically illustrated in Figure~\ref{fig:framework_schematic}), a framework that integrates function-space generative priors with differentiable neural operators to solve subsurface inverse problems.
We first formulate the Bayesian inverse problem (Section~\ref{sec:problem_formulation}), then present the function-space diffusion prior (Section~\ref{sec:prior}) and decoupled posterior sampling (Section~\ref{sec:fun-ddps}), and finally describe the CCS dataset and neural architectures (Section~\ref{sec:dataset_models}).
Table~\ref{tab:fun_ddps_vs_fundps} summarizes the key differences.

\begin{table}[b]
\centering
\begin{tabular}{lll}
\toprule
 & \textbf{Fun-DDPS} & \textbf{Fun-DPS} \\
\midrule
\textbf{Components} & Diffusion prior $p(\boldsymbol{m})$; & Joint diffusion prior $p(\boldsymbol{m}, \boldsymbol{s})$. \\
 & LNO surrogate $\mathcal{L}_\phi \approx F$. & \\
\textbf{Training data} & & \\
\quad Paired $(\boldsymbol{m}, \boldsymbol{s})$ & For surrogate only; fewer needed. & Required. \\
\quad Parameter only ($\boldsymbol{m}$) & For prior. & Not used. \\
\textbf{Output} & $\boldsymbol{m}$ from diffusion; $\boldsymbol{s} = \mathcal{L}_\phi(\boldsymbol{m})$. & $(\boldsymbol{m}, \boldsymbol{s})$ from joint model. \\
\textbf{Phys. enforcement} & Explicit. & Implicit. \\
\textbf{Guid. attenuation} & Avoided. & Occurs under data scarcity. \\
\bottomrule
\end{tabular}
\vspace{-0.5em}
\caption{Comparison of Fun-DDPS (decoupled) and Fun-DPS (joint-state).}
\label{tab:fun_ddps_vs_fundps}
\end{table}

\subsection{Problem Formulation}
\label{sec:problem_formulation}
Let $\boldsymbol{m} \in \mathcal{M}$ denote the static geological parameters (e.g., permeability) and $\boldsymbol{s} \in \mathcal{S}$ denote the dynamic state variables (e.g., saturation), defined on Hilbert spaces of functions.
The physics is governed by a forward operator $F: \mathcal{M} \to \mathcal{S}$.
We seek to infer the unknown geomodel $\boldsymbol{m}$ given sparse observations $\boldsymbol{y}_{obs}$.
The observation process is modeled as:
\begin{equation}
    \boldsymbol{y}_{obs} = \mathcal{H}(\boldsymbol{m}, F(\boldsymbol{m})) + \boldsymbol{\eta}, \quad \boldsymbol{\eta} \sim \mathcal{N}(0, \sigma_y^2 \mathbf{I})
\end{equation}
where $\mathcal{H}$ is a sparse sampling operator.
In the Bayesian setting, we target the posterior distribution:
\begin{equation}
    p(\boldsymbol{m} | \boldsymbol{y}_{obs}) \propto p(\boldsymbol{y}_{obs} | \boldsymbol{m}) p(\boldsymbol{m}),
\end{equation}
where $p(\boldsymbol{m})$ denotes the Radon--Nikodym derivative of the prior measure with respect to a Gaussian reference  measure on $\mathcal{M}$.
%
Our decoupled approach addresses this by learning a prior $p(\boldsymbol{m})$ independently and using a pre-trained neural operator surrogate to efficiently approximate the likelihood $p(\boldsymbol{y}_{obs} | \boldsymbol{m})$ during posterior sampling, allowing for robust posterior sampling even under extreme data sparsity.

\subsection{Function-Space Diffusion Prior}
\label{sec:prior}
We model the geological prior $p(\boldsymbol{m})$ using a Function-Space Diffusion Model~\citep{lim2023score}.

\textbf{Forward Process.}
Unlike finite-dimensional models, we define the forward process by perturbing $\boldsymbol{m}_0$ with Gaussian Random Fields (GRFs):
\begin{equation}
    \boldsymbol{m}_\sigma = \boldsymbol{m}_0 + \boldsymbol{\xi}_\sigma, \quad \boldsymbol{\xi}_\sigma \sim \mathcal{N}(0, \sigma^2 \mathcal{C}_\gamma)
\end{equation}
where $\mathcal{C}_\gamma$ is a covariance operator defined by a Matérn kernel and $\sigma$ is the noise level.
This ensures that the noise respects the spatial continuity of geological fields, preserving discretization invariance.
As $\sigma$ increases, the perturbed field $\boldsymbol{m}_\sigma$ gradually loses structure and approaches the base measure.

\textbf{Training Objective.}
We train a time-dependent denoising neural operator $\boldsymbol{D}_\theta(\boldsymbol{m}_\sigma, \sigma)$ to estimate the clean signal $\boldsymbol{m}_0$ from a noisy input $\boldsymbol{m}_\sigma$.
The network minimizes the score matching objective:
\begin{equation}
    \mathcal{L}(\theta) = \mathbb{E}_{\boldsymbol{m}_0, \sigma, \boldsymbol{\xi}} \left[ \lambda(\sigma) \| \boldsymbol{D}_\theta(\boldsymbol{m}_0 + \boldsymbol{\xi}_\sigma, \sigma) - \boldsymbol{m}_0 \|_{\mathcal{M}}^2 \right]
\end{equation}
where we choose $\|\cdot\|_{\mathcal{M}}$ to be $L_2$ norm, and $\lambda(\sigma)$ is a noise-level dependent weighting function.
Once trained, unconditional sampling from the prior $p(\boldsymbol{m})$ is performed via the probability flow ODE:
\begin{equation}
    d\boldsymbol{m}_\sigma = -\sigma \nabla_{\boldsymbol{m}} \log p_\sigma(\boldsymbol{m}_\sigma) \, d\sigma
\end{equation}
where $\nabla_{\boldsymbol{m}} \log p_\sigma(\boldsymbol{m}_\sigma)$ is the score function, approximated via $\boldsymbol{D}_\theta$.
We provide additional diagnostics of the learned geomodel prior (e.g., variograms and two-point statistics) in Appendix~\ref{app:prior_assessment}.

\subsection{Fun-DDPS: Decoupled Posterior Sampling}
\label{sec:fun-ddps}

\begin{figure}[t]
\centering
\includegraphics[clip,width=\columnwidth]{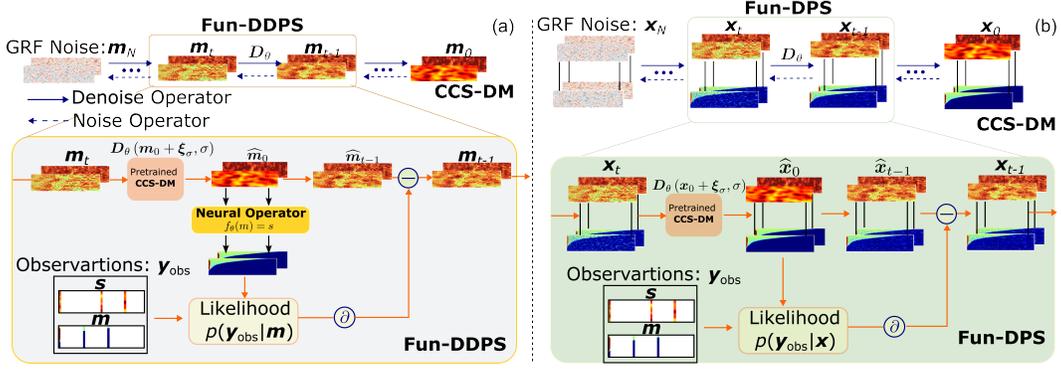}
\caption{Posterior sampling comparison: Fun-DDPS vs.\ Fun-DPS.
\textbf{(a) Fun-DDPS:} The geomodel diffusion model denoises a GRF to produce $\hat{\boldsymbol{m}}_0$. Geomodel observations provide direct guidance; dynamics observations guide via the surrogate gradient $\nabla_{\boldsymbol{m}} \|\mathcal{L}_\phi(\hat{\boldsymbol{m}}_0) - \boldsymbol{y}_{dyn}\|$, translating sparse solution-space constraints into dense parameter-space guidance.
\textbf{(b) Fun-DPS:} The joint-state model denoises to produce $(\hat{\boldsymbol{m}}_0, \hat{\boldsymbol{s}}_0)$ simultaneously. Observations guide the respective channels directly, but sparse dynamics observations remain localized without the surrogate's global receptive field to propagate information.}
\label{fig:GCS-DPS}
\end{figure}

To sample from the posterior, we follow Fun-DPS~\citep{yao2025guided}, which guides the reverse diffusion process in function space using the gradient of the log-likelihood.
Appendix~\ref{app:dps_details} provides unified pseudocode and implementation details for our posterior sampler.
The conditional score decomposes into the prior score (provided by $\boldsymbol{D}_\theta$) and the likelihood guidance:
\begin{equation}
    \nabla_{\boldsymbol{m}} \log p(\boldsymbol{m} | \boldsymbol{y}_{obs}) = \overbrace{\nabla_{\boldsymbol{m}} \log p(\boldsymbol{m})}^{\text{Prior (via } \boldsymbol{D}_\theta)} + \overbrace{\nabla_{\boldsymbol{m}} \log p(\boldsymbol{y}_{obs} | \boldsymbol{m})}^{\text{Likelihood guidance}},
\end{equation}
where $\nabla_{\boldsymbol{m}}$ denotes the Fr\'echet derivative of the corresponding functional.
The form of the likelihood guidance depends on what is observed.

\textbf{Forward modeling.}
Observations are of the geomodel itself, e.g.\ $\boldsymbol{y}_{geo} = M_{geo} \odot \boldsymbol{m} + \boldsymbol{\eta}$, so the likelihood gradient is computed directly and no surrogate is needed:
\begin{equation}
    \nabla_{\boldsymbol{m}} \log p(\boldsymbol{y}_{geo} | \hat{\boldsymbol{m}}_0) \approx -\zeta_{geo} \nabla_{\boldsymbol{m}} \| M_{geo} \odot (\hat{\boldsymbol{m}}_0 - \boldsymbol{y}_{geo}) \|_2^2
\end{equation}

\textbf{Inverse modeling.}
Observations are of the dynamic state, $\boldsymbol{y}_{dyn} = M_{dyn} \odot F(\boldsymbol{m}) + \boldsymbol{\eta}$, so computing the likelihood gradient requires differentiating through the forward map $F(\boldsymbol{m})$.
Since $F$ is expensive, we use a pre-trained neural operator surrogate $\mathcal{L}_\phi \approx F$ and backpropagate through $\mathcal{L}_\phi$:
\begin{equation}
    \nabla_{\boldsymbol{m}} \log p(\boldsymbol{y}_{dyn} | \hat{\boldsymbol{m}}_0) \approx -\zeta_{dyn} \nabla_{\boldsymbol{m}} \| M_{dyn} \odot (\mathcal{L}_\phi(\hat{\boldsymbol{m}}_0) - \boldsymbol{y}_{dyn}) \|_2^2
    \label{eq:dyn_guidance}
\end{equation}

Eq.~\ref{eq:dyn_guidance} represents the core innovation of Fun-DDPS.
By separating the prior $p(\boldsymbol{m})$ from the physics $\mathcal{L}_\phi$, generated samples remain on the learned geological manifold while data consistency is enforced through surrogate guidance.
This mechanism is visualized in Figure~\ref{fig:GCS-DPS}, which highlights how Fun-DDPS translates sparse constraints in the solution space into dense guidance in the parameter space via the surrogate's Jacobian.
This contrasts with some joint-state approaches (e.g. Fun-DAPS in \citet{lin2025ddis}), where gradients from sparse observations remain localized to the specific channels, often resulting in high-frequency artifacts and physical inconsistencies when training data is limited.

\paragraph{Theoretical Motivation.}
Recent analysis by~\citet{lin2025ddis} shows that joint-state models suffer from \textit{guidance attenuation} under data scarcity: the gradient signal that updates coefficients vanishes when training data is limited.
This occurs because effective guidance requires the diffusion state to lie near multiple training samples simultaneously, a condition that becomes exponentially rare in high dimensions.
In contrast, the decoupled design avoids this failure by using the explicit neural operator Jacobian $\nabla_{\boldsymbol{m}} \mathcal{L}_\phi(\boldsymbol{m})$ for guidance, which does not depend on training data density.
%

\subsection{CCS Dataset and Neural Architectures}
\label{sec:dataset_models}

\subsubsection{Dataset Generation}
To validate our framework, we simulate the injection of supercritical CO$_2$ into a radially symmetrical deep saline aquifer.
We generate a dataset of 12,000 training pairs and 1,500 test pairs using the industry-standard simulator ECLIPSE (e300)~\citep{eclipse}.
Each realization consists of a heterogeneous permeability field $\boldsymbol{m}$ generated via SGeMS~\citep{sgems} and the corresponding CO$_2$ saturation field $\boldsymbol{s}$ after 30 years of injection.
The governing multiphase flow equations, numerical grid settings, and prior distributions for geological parameters are detailed in Appendix~\ref{app:pde_training_details}.

\subsubsection{Surrogate Architecture (LocalNO)}
For the forward surrogate $\mathcal{L}_\phi$, we employ a Local Neural Operator (LNO)~\citep{liu2024neural} to learn the mapping from the geomodel $\boldsymbol{m}$ to the dynamic state $\boldsymbol{s}$.
The training set consists of 12,000 pairs $\{(\boldsymbol{m}_i, \boldsymbol{s}_i)\}$.
While standard operators such as FNO~\citep{li2020fourier} excel at capturing global dependencies, they often produce ringing artifacts around sharp discontinuities. 
The LNO addresses this by combining global Fourier spectral layers with localized kernels implemented via Discrete Continuous Convolutions (DISCO).
This dual-path design is critical for our application: the Fourier path resolves global pressure responses, while the DISCO path accurately captures the shock-like saturation fronts characteristic of multiphase flow.
We note that the surrogate is relatively robust to train even when paired data are limited, and our sampling pipeline is tolerant to surrogate approximation error.

\subsubsection{Diffusion Backbone (U-NO)}
Both the decoupled geomodel prior (Fun-DDPS) and the joint-state baseline (Fun-DPS) utilize a U-shaped Neural Operator (U-NO) architecture~\citep{rahman2022u, yao2025guided}.
The U-NO integrates neural operator blocks into a multi-scale U-Net structure, ensuring discretization invariance while capturing features at varying spatial resolutions.
The same 12,000 training samples are used to train both diffusion models, though the U-NO in Fun-DDPS is trained solely on the geomodel fields $\{\boldsymbol{m}_i\}$, whereas Fun-DPS is trained on the joint input--output pairs.
To ensure a rigorous comparison, we fix the architectural hyperparameters to be identical for both methods (See Appendix~\ref{app:training_details}). We also examine the quality of the learned geomodel priors both qualitatively and quantitatively in Appendix~\ref{app:prior_assessment}.

\section{Experiments}
\label{sec:experiments}

We evaluate the proposed Fun-DDPS framework on two distinct tasks: (1) forward modeling conditioned on partial static inputs (geomodel), and (2) inverse modeling conditioned on partial dynamic observations (saturation), as schematically shown in Fig.~\ref{fig:forward_inverse_setup}.
In practical deployment, the observation sparsity ratio is usually unknown a priori and can vary across settings.
To reflect this, our training pipeline is agnostic to the observation-mask ratio, and sparsity is imposed only at inference time for evaluation.
For the forward problem, we benchmark performance against two baselines: \textbf{Fun-DPS:} A joint-state diffusion baseline; and \textbf{Surrogate ($\mathcal{L}_\phi$):} A deterministic neural operator trained on complete inputs. For the inverse problem, we provide rigorous validation by comparing our approximate posteriors against a ``ground truth'' posterior distribution obtained via Rejection Sampling (RS)~\citep{robert1999monte}.

\subsection{Forward Problem: Partial Geomodel Observations}
\label{sec:forward}

\subsubsection{Problem Setup}
\label{sec:forward_setup}

\begin{figure}[t]
    \centering
    \includegraphics[width=0.85\columnwidth]{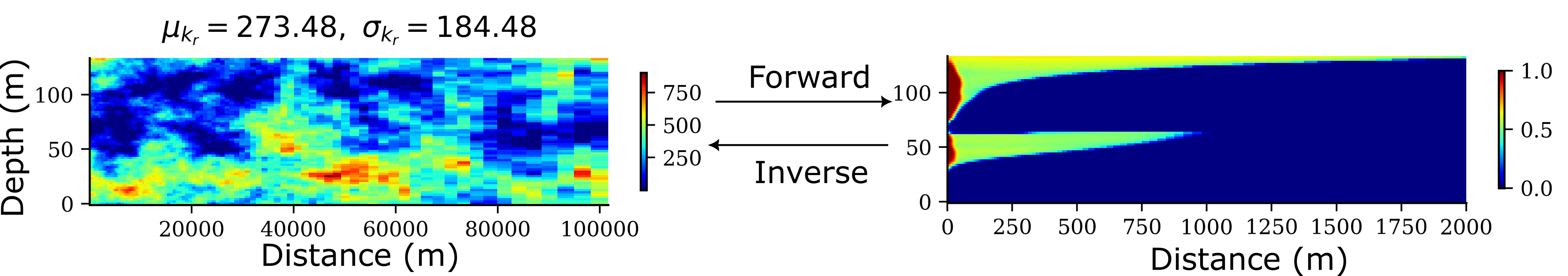}
    \caption{Problem setup. \textbf{Forward:} The geomodel is partially observed; the full dynamics are predicted. \textbf{Inverse:} The dynamics are partially observed; the full geomodel is inferred.}
    \label{fig:forward_inverse_setup}
\end{figure}

In this task, the goal is to predict the dynamic saturation field $\boldsymbol{s}$ given a partially observed geomodel $\boldsymbol{y}_{geo}$. We evaluate performance across three observation sparsity levels: 100\% (full), 50\% (random), and 25\% (random) coverage.
The forward/inverse task setup is shown in Figure~\ref{fig:forward_inverse_setup}.

The primary challenge here is that deterministic surrogates ($\mathcal{L}_\phi$) require dense, complete inputs. Standard approaches, such as zero-filling, push the input out of the training distribution, typically leading to severe prediction errors. Fun-DDPS addresses this by utilizing the learned diffusion prior to probabilistically reconstruct missing geomodel features \textit{before} the forward simulation.

For evaluation, we use a test set of 480 samples. We run the diffusion process for 500 iterations with a guidance weight of $\zeta_{geo} = 10{,}000$ to ensure strong consistency with the observed data points.

\subsubsection{Results}
\label{sec:forward_result}

Table~\ref{tab:forward_results} summarizes the relative $L_2$ errors. At full observation (100\%), Fun-DDPS methods perform comparably to the neural surrogate. However, as sparsity increases, the deterministic surrogate degrades catastrophically---rising from 4.4\% to 86.9\% error---confirming that zero-filling is insufficient for complex physical mappings.
In contrast, Fun-DDPS exhibits remarkable robustness, maintaining a low 7.7\% error even at 25\% coverage. This represents an 11$\times$ improvement over the surrogate. By leveraging the generative prior, Fun-DDPS reconstructs a physically plausible full-field geomodel, allowing the underlying operator to function within its valid input domain.

\begin{table}[b]
\centering
\small
\begin{tabular}{lccc}
\toprule
Obs.\ Coverage & Fun-DDPS & Fun-DPS & Surrogate \\
\midrule
100\% & 0.046 $\pm$ 0.058 & 0.418 $\pm$ 0.366 & \textbf{0.044 $\pm$ 0.057} \\
50\% & \textbf{0.054 $\pm$ 0.069} & 0.370 $\pm$ 0.357 & 0.850 $\pm$ 0.263 \\
25\% & \textbf{0.077 $\pm$ 0.079} & 0.336 $\pm$ 0.349 & 0.869 $\pm$ 0.258 \\
\bottomrule
\end{tabular}
\caption{Forward problem: Relative $L_2$ error (mean $\pm$ std). Fun-DDPS maintains robustness under high sparsity, whereas the deterministic surrogate fails due to input artifacts.}
\label{tab:forward_results}
\vspace{-1em}
\end{table}

\begin{figure}[t]
    \centering
    \includegraphics[width=\columnwidth]{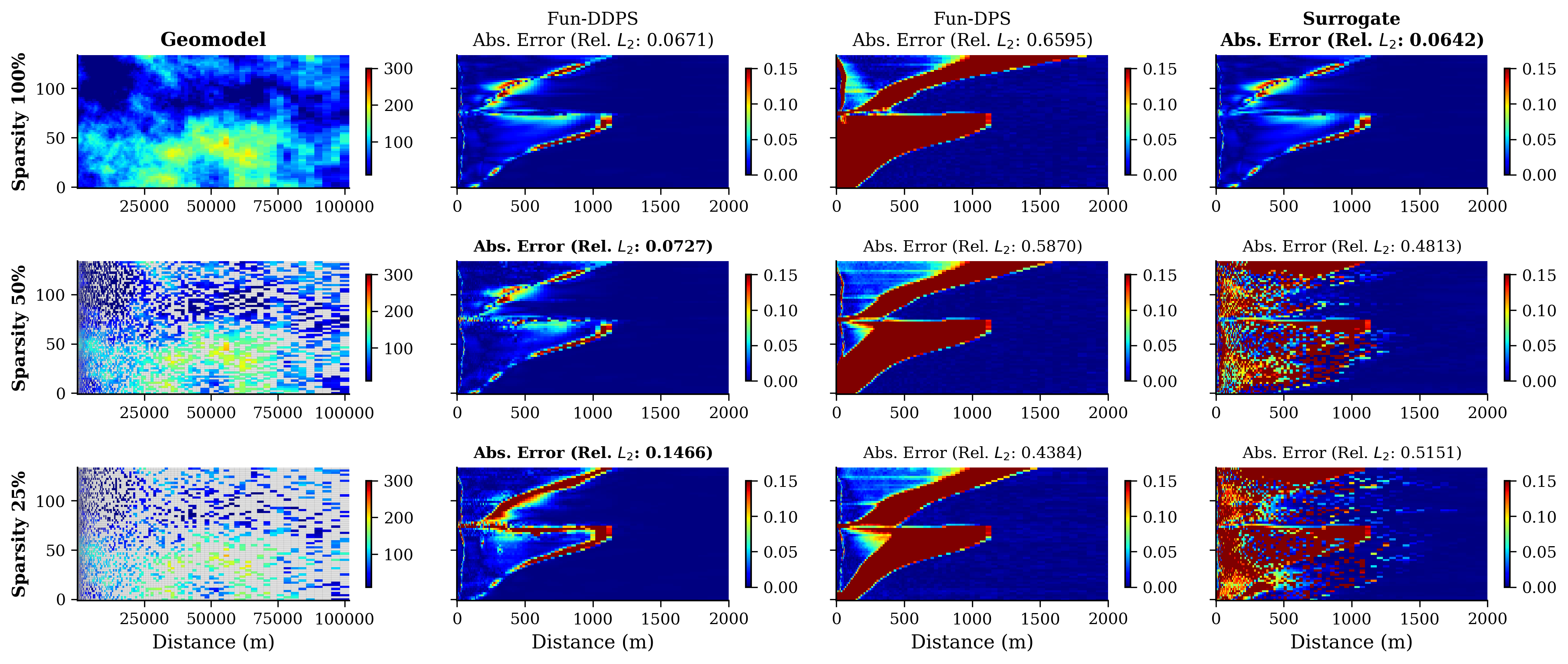}
    \caption{Forward problem under varying observation coverage. \textbf{Left:} Observed geomodel at 100\%, 50\%, and 25\% coverage. \textbf{Right:} Absolute error in predicted saturation. Fun-DDPS maintains high accuracy (rel.\ $L_2$: 0.07--0.15).}
    \label{fig:forward_results}
\end{figure}

Qualitative results are visualized in Figure~\ref{fig:forward_results}. While the surrogate produces noisy, incoherent predictions in the presence of gaps, Fun-DDPS maintains coherent spatial structures, particularly capturing the sharp gradients at the CO$_2$ plume front. 
Notably, the Fun-DPS baseline shows consistently high error (34--42\%) regardless of sparsity.
This performance gap highlights a fundamental advantage of the proposed decoupled architecture over joint-state diffusion models.
While joint-state models primarily encode geomodel--saturation coupling as statistical correlation in the joint distribution and enforce PDE only at inference time via a physics loss, they can both miss physical laws under limited supervision and have higher sampling cost from physics-loss evaluations.
In contrast, Fun-DDPS decouples the generation and simulation processes: it employs a pre-trained neural operator to strictly enforce physical consistency.
Because the operator is trained exclusively on the mapping task, it learns to honor the underlying physics much more effectively than a generative model attempting to learn both the prior and the forward map simultaneously.
This observation aligns with findings in DDIS~\cite{lin2025ddis}, which show that decoupled methods often achieve superior sampling efficiency and physical fidelity compared to joint-state approaches when paired training data is scarce.

\subsection{Inverse Problem: Partial Dynamics Observations}
\label{sec:inverse}

In this section, we address the inverse problem: conditioning on sparse dynamic observations $\boldsymbol{y}_{dyn}$ to infer the underlying geomodel $\boldsymbol{m}$.

\subsubsection{Problem Setup}
\label{sec:inverse_setup}

\begin{figure}[t]
\centering
\begin{minipage}[c]{0.32\columnwidth}
\centering
\small
\vspace{-0.5em}
\begin{tabular}{lcc}
\toprule
 & $\mu_{k_r}$ (mD) & $\sigma_{k_r}$ (mD) \\
\midrule
Prior & $\mathcal{U}[10, 500]$ & $\mathcal{U}[1, 500]$ \\
True & 223.0 & 41.7 \\
\bottomrule
\end{tabular}
\captionof{table}{Geomodel hyperparameters for the ground truth case and their corresponding prior ranges.}
\label{tab:prior_true_params}
\end{minipage}
\hfill
\begin{minipage}[c]{0.65\columnwidth}
\centering
\vspace{-3em}
\includegraphics[width=\textwidth]{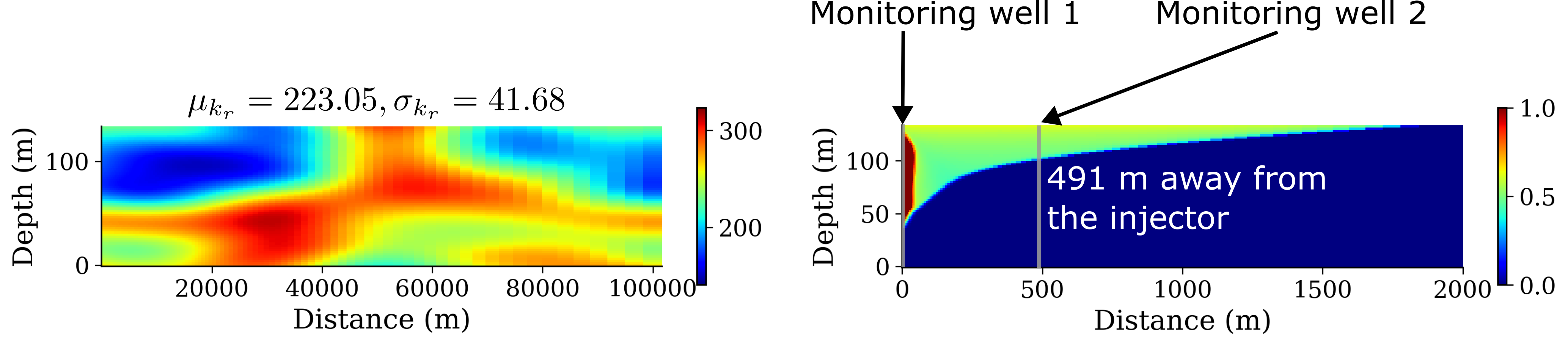}
\captionof{figure}{Inverse problem setup. Ground truth permeability ($k_r$) and gas saturation ($s_g$) after 30 years. Vertical lines indicate the two monitoring wells where sparse saturation data is assimilated.}
\label{fig:da_setup}
\end{minipage}
\end{figure}

\paragraph{Physical Configuration \& Ground Truth}
The ground truth geomodel is generated using SGeMS \citep{remy2009applied}, with permeability hyperparameters $\mu_{k_r}$ and $\sigma_{k_r}$ sampled from uniform priors (See Table~\ref{tab:prior_true_params}). The saturation field $s_g$ is simulated over a 30-year injection period using a high-fidelity numerical solver.

We design the experimental setup to reflect the extreme data sparsity encountered in practical Carbon Capture and Storage (CCS) projects. 
In real-world scenarios, subsurface data is often restricted to well logs, which provide only 1D vertical profiles (column observations) within a vast domain. To mimic this challenge, observations are collected from just two monitoring wells: one at the injector and one 491 meters away (Figure~\ref{fig:da_setup}).
Each well provides 64 dynamic data points, totaling 128 measurements. This configuration corresponds to observing less than 1\% of the spatial domain, imposing an extreme challenge for inverse methods to resolve the heterogeneous geomodel from such limited information. We add Gaussian noise with $\sigma_{obs} = 0.04$ to the observations.

\paragraph{Reference Posterior by Rejection Sampling}
To rigorously validate the diffusion-based posteriors, we compute a ``ground truth'' reference posterior using Rejection Sampling (RS). We draw from a pool of 2 million prior samples, accepting samples based on the likelihood:
$$
L(\boldsymbol{m}) = \exp\left(-\frac{\|\mathbf{M} \odot (\mathcal{L}_\phi(\boldsymbol{m}) - \boldsymbol{y}_{dyn})\|^2}{2\sigma_{obs}^2 \cdot |\mathbf{M}|}\right)
$$
This yields approximately 26,000 accepted samples, providing a dense representation of the true posterior distribution. The details of the RS algorithm are provided in Appendix~\ref{app:rs_benchmark}.

\paragraph{Inference Setup}
We generate 1,024 posterior samples using both Fun-DDPS and Fun-DPS. To ensure a fair comparison, we performed a hyperparameter sweep to determine the optimal guidance strength $\zeta_{dyn}$ for each method, effectively weighting the data-consistency term relative to the observation noise. 
Based on this sweep, the guidance strengths for Fun-DDPS and Fun-DPS are set to $\zeta_{dyn} = 45.0$ and $\zeta_{dyn} = 100.0$, respectively.

\subsubsection{Results}
\label{sec:inverse_result}

\begin{table}[t]
\centering
\small
\begin{tabular}{lcccc}
\toprule
Method & JS (Mean) & JS (Std) & $N$ & Filter \\
\midrule
RS (Ref.) & -- & -- & 26,082 & -- \\
Fun-DPS & \textbf{0.047} & \textbf{0.037} & 1,024 & 0\% \\
Fun-DDPS & 0.051 & 0.061 & 998 & 2.5\% \\
\bottomrule
\end{tabular}
\caption{
    Inverse problem validation. JS divergence measures the distance to the RS reference posterior. Both methods achieve high accuracy (JS $< 0.06$). The Filter column indicates the percentage of diverged samples ($k > 5000$\,mD).
    We note that despite slightly worse JS scores, Fun-DDPS produces more physically plausible geomodels as seen in Figure~\ref{fig:posterior_qualities}.
}
\label{tab:inverse_results}
\end{table}

\begin{figure}[t]
    \centering
    \includegraphics[width=0.9\columnwidth]{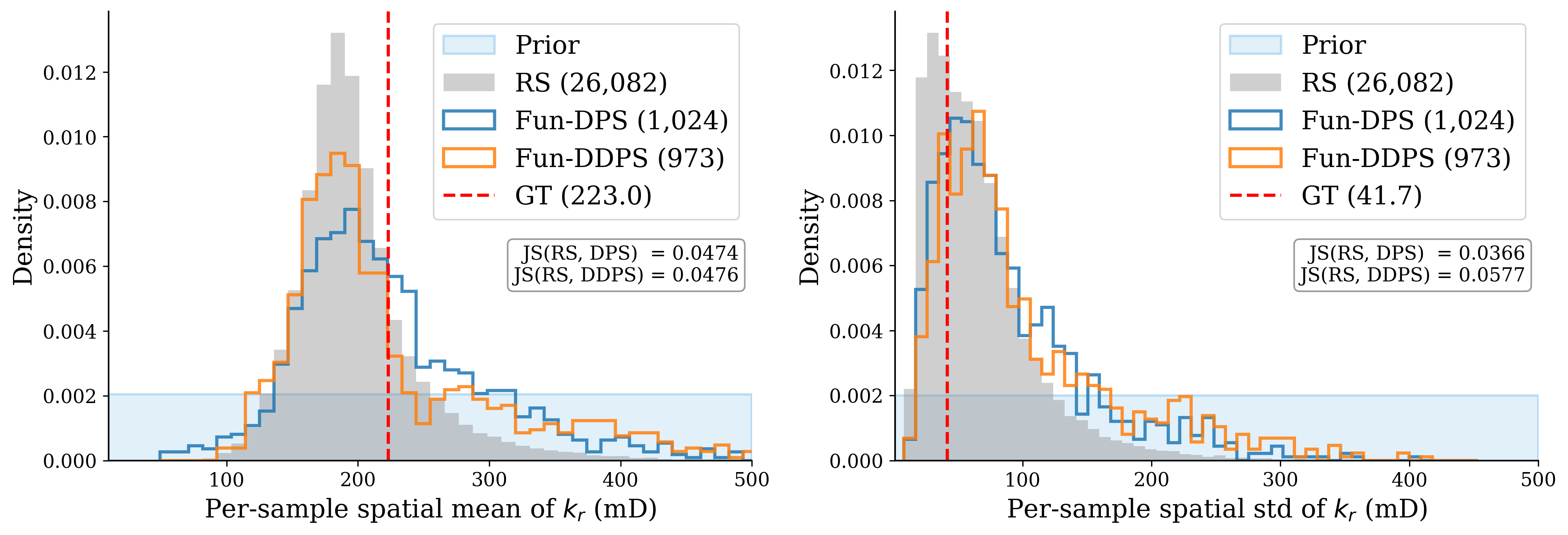}
    \caption{Marginal posterior distributions of geomodel hyperparameters ($\mu_{k_r}$, $\sigma_{k_r}$) comparing RS (gray), Fun-DPS (blue), and Fun-DDPS (orange), overlaid on the uniform prior (light blue shaded region). Vertical dashed lines indicate true values.}
    \label{fig:posterior_rs_ddps}
\end{figure}

Figure~\ref{fig:posterior_rs_ddps} compares the marginal posterior distributions of geomodel hyperparameters ($\mu_{k_r}$, $\sigma_{k_r}$) obtained from RS, Fun-DPS, and Fun-DDPS, overlaid on the uniform prior distributions (Table~\ref{tab:prior_true_params}).
Moreover, all three posteriors are sharply concentrated relative to the broad prior, demonstrating significant uncertainty reduction from the measured dynamic data.
The RS posterior serves as the ground-truth reference, capturing the true parameter values within its support.
Both diffusion methods successfully recover the posterior mode location: Fun-DPS (blue) shows tighter agreement with the RS distribution shape, while Fun-DDPS (orange) exhibits slightly broader tails.

\begin{figure}[t]
    \centering
    \includegraphics[width=\columnwidth]{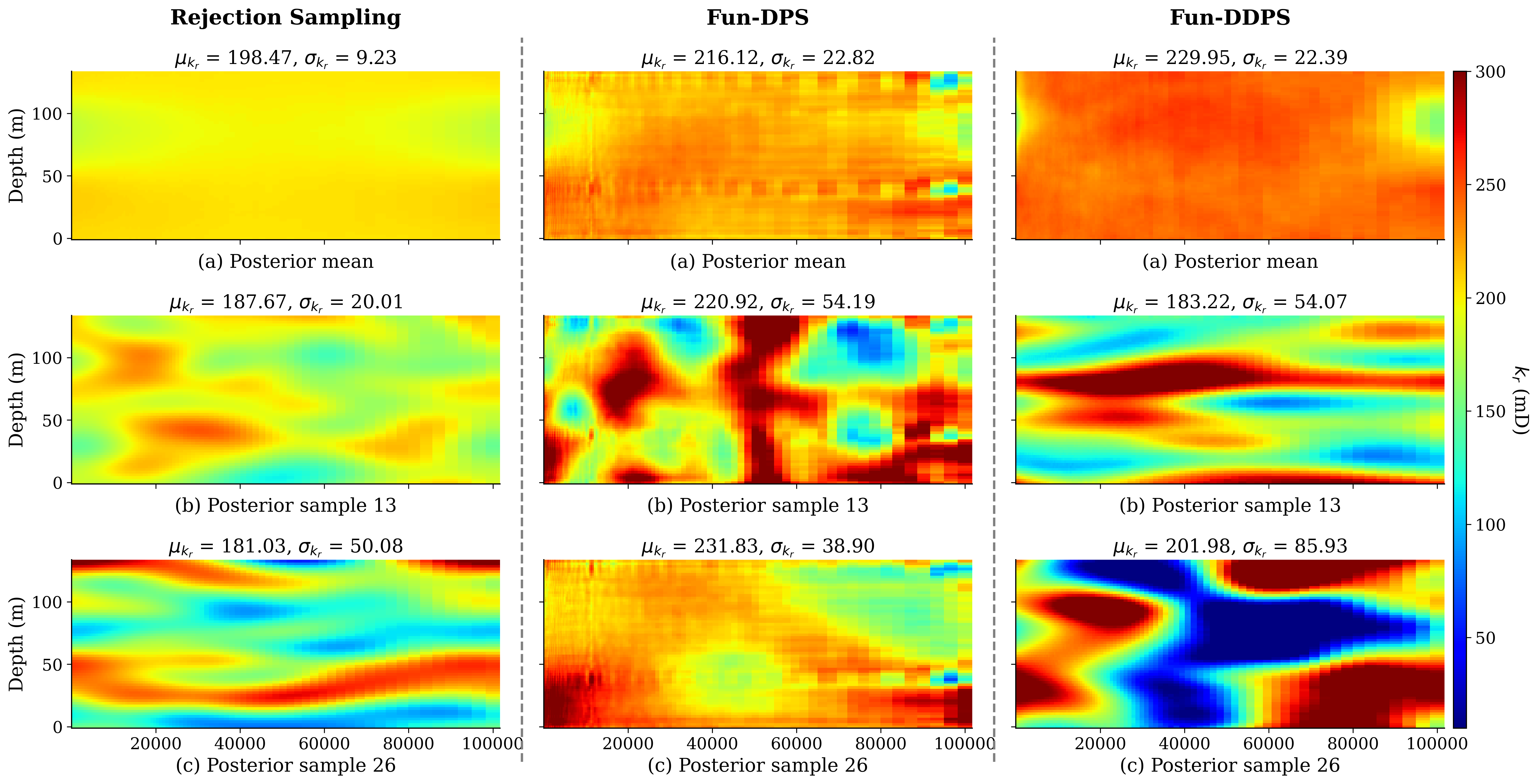}
    \caption{Inverse problem: posterior comparison. Each column shows posterior mean (row a) and two samples (rows b, c). \textbf{RS} (left): ground-truth with 26,082 samples. \textbf{Fun-DPS} (center): captures posterior structure but suffers from obvious high-frequency artifacts, visible in the grainy posterior mean. \textbf{Fun-DDPS} (right): generates more physically plausible geomodels with smoother spatial features, avoiding the artifacts seen in the joint-state baseline.}
    \label{fig:posterior_qualities}
\end{figure}

However, a visual inspection of the spatial fields in Figure~\ref{fig:posterior_qualities} reveals a critical qualitative difference between the methods.
While Fun-DPS is statistically precise, its generated samples tend to exhibit non-physical, high-frequency artifacts. This noise is particularly evident in the Fun-DPS posterior mean (Fig.~\ref{fig:posterior_qualities}, center column), which appears grainy and textured compared to the smooth reference mean.
In contrast, Fun-DDPS produces geologically coherent realizations that maintain physical continuity. By decoupling the generation process, Fun-DDPS effectively leverages the prior to suppress high-frequency noise, resulting in a posterior mean that is cleaner and visually closer to the smoothness of the RS reference.

Finally, in terms of computational efficiency, Fun-DDPS demonstrates a significant advantage.
Generating 1,024 posterior samples required approximately 512,000 functional evaluations ($1,024 \text{ samples} \times 500 \text{ steps}$), whereas obtaining the reference RS posterior required 2 million evaluations.
This represents an approximately $4\times$ reduction in the computational budget.

\section{Conclusion}
\label{sec:conl}

We presented Fun-DDPS, a unified framework for geological CO$_2$ storage that decouples the learning of complex geological priors from the approximation of subsurface flow physics.
Our approach addresses the dual challenges of extreme data sparsity and computational expense in subsurface data assimilation.

We demonstrate three critical advantages over existing methods.
First, Fun-DDPS exhibits exceptional robustness in forward modeling with partial inputs: while standard deterministic surrogates fail catastrophically (87\% error) when facing 25\% data coverage, Fun-DDPS leverages its generative prior to achieve a relative error of only 7.7\%.
Second, we provided the first rigorous validation of diffusion-based inverse solvers against asymptotically exact Rejection Sampling (RS) posteriors.
We showed that while both decoupled and joint-state models achieve high statistical accuracy (Jensen-Shannon divergence $<0.06$), the decoupled approach produces significantly more physically consistent realizations free from the high-frequency artifacts observed in joint-state baselines.
Finally, this performance is achieved with high efficiency, approximating the true posterior with a $4\times$ reduction in the total computational budget compared to Rejection Sampling.

While Fun-DDPS demonstrates strong performance on the presented CCS tasks, our current study simplifies the dynamic state $\boldsymbol{s}$ to a single temporal snapshot (30 years post-injection).
In realistic CO$_2$ monitoring scenarios, data assimilation involves continuous streams of time-series data (e.g., daily pressure gauges, 4D seismic surveys) rather than isolated snapshots.
Extending our framework to handle full spatiotemporal trajectories is a natural next step. The decoupled architecture is well-suited for this: the neural operator can map static geomodels to time-varying dynamic states without retraining the expensive diffusion prior.




\subsubsection*{Acknowledgments}
Anima Anandkumar is supported in part by Bren endowed chair, ONR (MURI grant N00014-23-1-2654), and the AI2050 senior fellow program at Schmidt Sciences.
Gege Wen acknowledges the generous support of Schmidt Sciences through the AI2050 fellowship.

\bibliography{ref}
\bibliographystyle{iclr2026_conference}
\newpage

\appendix

\include{appendix}

\end{document}

%% file: math_commands.tex
\usepackage{amsmath,amsfonts,bm}

\def\eqref#1{equation~\ref{#1}}

\def\1{\bm{1}}

\DeclareMathAlphabet{\mathsfit}{\encodingdefault}{\sfdefault}{m}{sl}
\SetMathAlphabet{\mathsfit}{bold}{\encodingdefault}{\sfdefault}{bx}{n}

%% file: appendix.tex
\section{Diffusion Posterior Sampling (DPS)}
\label{app:dps_details}

Here we provide the unified algorithm for posterior sampling (Algorithm~\ref{alg:posterior_sampling}). We formulate the generic algorithm for state variable $\boldsymbol{z}$, which corresponds to $\boldsymbol{m}$ in Fun-DDPS and $\boldsymbol{x}=(\boldsymbol{m},\boldsymbol{s})$ in Fun-DPS.
The following details are presented in a finite-dimensional setting; for the function-space variant, please refer to Fun-DPS~\citep{yao2025guided}.

To evaluate the conditional score $\nabla_{\boldsymbol{z}} \log p(\boldsymbol{y}_{obs} | \boldsymbol{z}_\sigma)$, we employ the standard DPS approximation~\citep{chung2022diffusion} using the denoised estimate $\hat{\boldsymbol{z}}_0$. 
By Tweedie's formula, we have:
\begin{equation}
    \hat{\boldsymbol{z}}_0(\boldsymbol{z}_\sigma) = \boldsymbol{D}_\theta(\boldsymbol{z}_\sigma, \sigma) \approx \boldsymbol{z}_\sigma + \sigma^2 \nabla_{\boldsymbol{z}} \log p(\boldsymbol{z}_\sigma)
\end{equation}
The conditional score (a.k.a.the observation guidance gradient) is then approximated via Maximum Likelihood Estimation (MLE):
\begin{equation}
    \nabla_{\boldsymbol{z}_\sigma} \log p(\boldsymbol{y}_{obs} | \boldsymbol{z}_\sigma) \approx
    \nabla_{\boldsymbol{z}_\sigma} \log p(\boldsymbol{y}_{obs} | \hat{\boldsymbol{z}}_0(\boldsymbol{z}_\sigma))
    = -\zeta \nabla_{\boldsymbol{z}_\sigma} \| \boldsymbol{y}_{obs} - \mathcal{H}(\hat{\boldsymbol{z}}_0(\boldsymbol{z}_\sigma)) \|_2^2
\label{eq:app_guidance_grad}
\end{equation}

\begin{algorithm}[H]
\caption{Unified Diffusion Posterior Sampler}
\label{alg:posterior_sampling}
\begin{algorithmic}[1]
\Require Observations $\boldsymbol{y}_{obs}$, trained denoiser $\boldsymbol{D}_{\theta}$, noise schedule $\{\sigma_i\}_{i=0}^N$ (descending), guidance weight $\zeta$, GRF covariance $\mathbf{C}_{\gamma}$.
\State \textbf{Setup:}
\If{Method is \textbf{Fun-DDPS}}
    \State State $\boldsymbol{z} \leftarrow \boldsymbol{m}$ \Comment{Sampling Geomodel only}
    \State Operator $\mathcal{H}_{obs}(\hat{\boldsymbol{z}}_0) \leftarrow M_{dyn} \odot \mathcal{L}_\phi(\hat{\boldsymbol{m}}_0)$ \Comment{Backprop through Surrogate $\mathcal{L}_\phi$}
\ElsIf{Method is \textbf{Fun-DPS}}
    \State State $\boldsymbol{z} \leftarrow (\boldsymbol{m}, \boldsymbol{s})$ \Comment{Sampling Joint State}
    \State Operator $\mathcal{H}_{obs}(\hat{\boldsymbol{z}}_0) \leftarrow M_{dyn} \odot \hat{\boldsymbol{s}}_0$ \Comment{Direct channel guidance}
\EndIf
\State $\boldsymbol{z}_N \sim \mathcal{N}(0, \sigma_N^2 \mathbf{C}_{\gamma})$ \Comment{Initialize from prior}

\State \textbf{Loop:}
\For{$i = N$ \textbf{to} $1$}
    \State $\hat{\boldsymbol{z}}_0^{(i)} \leftarrow \boldsymbol{D}_{\theta}(\boldsymbol{z}_i, \sigma_i)$ \Comment{Estimate clean state (Tweedie's formula)}
    \State $\boldsymbol{d}_i \leftarrow (\boldsymbol{z}_i - \hat{\boldsymbol{z}}_0^{(i)})/\sigma_i$ \Comment{Compute score/direction}
    
    \State \textcolor{gray}{// 1. Predictor Step (Euler's method)}
    \State $\boldsymbol{z}_{i-1}^{pred} \leftarrow \boldsymbol{z}_i + (\sigma_{i-1} - \sigma_i) \boldsymbol{d}_i$

    \State \textcolor{gray}{// 2. Corrector Step (2nd-order Heun's method)}
    \If{$i > 1$ \textbf{and} $\sigma_{i-1} > 0$}
        \State $\hat{\boldsymbol{z}}_0^{pred} \leftarrow \boldsymbol{D}_{\theta}(\boldsymbol{z}_{i-1}^{pred}, \sigma_{i-1})$
        \State $\boldsymbol{d}_{i-1} \leftarrow (\boldsymbol{z}_{i-1}^{pred} - \hat{\boldsymbol{z}}_0^{pred}) / \sigma_{i-1}$ \Comment{Direction at next step}
        \State $\boldsymbol{z}_{i-1} \leftarrow \boldsymbol{z}_i + (\sigma_{i-1} - \sigma_i) \left( \frac{1}{2}\boldsymbol{d}_i + \frac{1}{2}\boldsymbol{d}_{i-1} \right)$ \Comment{Trapezoidal update}
    \Else
        \State $\boldsymbol{z}_{i-1} \leftarrow \boldsymbol{z}_{i-1}^{pred}$
    \EndIf

    \State \textcolor{gray}{// 3. Guidance Step}
    \If{$\sigma_{i-1} > 0$}
        \State $\hat{\boldsymbol{z}}'_{0} \leftarrow \boldsymbol{D}_{\theta}(\boldsymbol{z}_{i-1}, \sigma_{i-1})$ \Comment{Denoise for gradient calculation}
        \State $\boldsymbol{g} \leftarrow \nabla_{\boldsymbol{z}_{i-1}} \| \boldsymbol{y}_{obs} - \mathcal{H}_{obs}(\hat{\boldsymbol{z}}'_{0}) \|_2^2$
        \State $\boldsymbol{z}_{i-1} \leftarrow \boldsymbol{z}_{i-1} - \zeta \cdot \boldsymbol{g}$ \Comment{Apply gradient guidance}
    \EndIf
\EndFor
\State \textbf{return} $\boldsymbol{D}_{\theta}(\boldsymbol{z}_0, \sigma_0)$
\end{algorithmic}
\end{algorithm}

\section{Joint-State Diffusion Models}
\label{app:joint_state_dm}

This section details the joint-state diffusion model, with DiffusionPDE~\citep{huang2024diffusionpde} and Fun-DPS~\citep{yao2025guided} as the two representative examples. While the proposed Fun-DDPS (Section~\ref{sec:methodology}) decouples the prior $p(\boldsymbol{m})$ from the physics, the joint-state approach learns the joint distribution $p(\boldsymbol{m}, \boldsymbol{s})$ directly.

We first define the joint state $\boldsymbol{x} = (\boldsymbol{m}, \boldsymbol{s}) \in \mathcal{X}$, where $\mathcal{X} = \mathcal{M} \times \mathcal{S}$ is the product Hilbert space of parameters and states. The diffusion model is trained on pairs $\{\boldsymbol{x}^{(i)}\}_{i=1}^N$ generated by the simulator, implicitly capturing physical constraints within the joint distribution.

We introduce a centered Gaussian reference measure $\mathcal{N}(0, \mathbf{C}_\mathcal{X})$ on the joint space. The forward process perturbs a clean sample $\boldsymbol{x}_0$ via:
\begin{equation}
    \boldsymbol{x}_\sigma = \boldsymbol{x}_0 + \boldsymbol{\xi}_\sigma, \quad \boldsymbol{\xi}_\sigma \sim \mathcal{N}(0, \sigma(t)^2 \mathbf{C}_\mathcal{X})
\label{eq:app_noising_process}
\end{equation}
where $\sigma(t)$ is the noise schedule.

We train a joint denoising operator $\boldsymbol{D}_\theta(\boldsymbol{x}_{\sigma}, \sigma)$ to recover $\boldsymbol{x}_0$. The loss function is the function-space analogue of denoising score matching:
\begin{equation}
    \mathcal{L}_{joint}(\theta) = \mathbb{E}_{\boldsymbol{x}_0, \sigma, \boldsymbol{\xi}_\sigma} \left[ \lambda(\sigma) \| \boldsymbol{D}_\theta(\boldsymbol{x}_0 + \boldsymbol{\xi}_\sigma, \sigma) - \boldsymbol{x}_0 \|_{\mathcal{X}}^2 \right]
\label{eq:app_denoising_loss}
\end{equation}
It is worth noting that during training, the physics is learned implicitly through the cross-state statistics and is not part of the training objective.

\section{Rejection Sampling Benchmark}
\label{app:rs_benchmark}

This section details the Rejection Sampling (RS) procedure used to obtain ground-truth posteriors for validating Fun-DDPS and Fun-DPS in Section~\ref{sec:inverse}.

\subsection{Motivation}

RS provides exact posterior samples when proposals are drawn from the prior. Unlike diffusion-based methods that approximate the posterior, RS serves as a rigorous benchmark. However, RS requires many forward evaluations due to low acceptance rates; we use a pre-trained neural operator surrogate $\mathcal{L}_\phi$ to make this computationally tractable.

\subsection{Problem Setup}
We use the same setup as the one shown in Fig.~\ref{fig:da_setup}, where two columns of gas saturation are observed. 
We use the identical ground truth as the one described in the previous section, where the same observation models and noise variance are used for conducting RS. 
As the RS method is essentially a rigorous search method that characterizes the exact posterior distribution, we need to generate a large number of prior samples, as the acceptance rate can be very low. 
Specifically, for generating prior geomodels, we first uniformly sample hyperparameters and later use SGeMS to generate 2 million geomodels that cover the prior space. Then, the pretrained surrogate model is used as the forward model to predict the $s_g$ distribution. 

To quantitatively measure the discrepancy between the posterior distributions and the RS ground truth, we use the Jensen-Shannon (JS) divergence.
The JS divergence is a symmetric and smoothed version of the Kullback-Leibler (KL) divergence~\cite{bishop2006pattern}, providing a bounded metric for the similarity between two probability distributions. It is defined as:
\begin{equation}
\mathrm{JS}(P \| Q) = \frac{1}{2} D_{KL}(P \| M) + \frac{1}{2} D_{KL}(Q \| M)
\end{equation}
where $P$ is the approximate posterior distribution (from machine learning models), $Q$ is the reference posterior (from RS), and $M = \frac{1}{2}(P+Q)$ is the average of the two distributions. A JS divergence of 0 indicates that the distributions are identical, while a larger value signifies a greater dissimilarity.

\section{Prior Assessment of Fun-DDPS}
\label{app:prior_assessment}

A critical step in validating the Fun-DDPS framework is ensuring that the decoupled generation process produces statistically and physically representative samples.
Unlike joint-state models, Fun-DDPS generates samples in a two-step pipeline: first, unconditional geomodels are drawn from the learned diffusion prior $\boldsymbol{m} \sim p_\theta(\boldsymbol{m})$; second, these realizations are mapped through the pre-trained LocalNO surrogate to obtain the corresponding dynamic states $\boldsymbol{s} = \mathcal{L}_\phi(\boldsymbol{m})$.
To assess this composite pipeline, we generated an ensemble of 1,000 unconditional pairs $\{(\boldsymbol{m}^{(i)}, \boldsymbol{s}^{(i)})\}$ and compared them against a ground-truth test set of equal size.

\subsection{Qualitative Visual Inspection}

\begin{figure}[H]
    \centering
    \includegraphics[width=0.8\columnwidth]{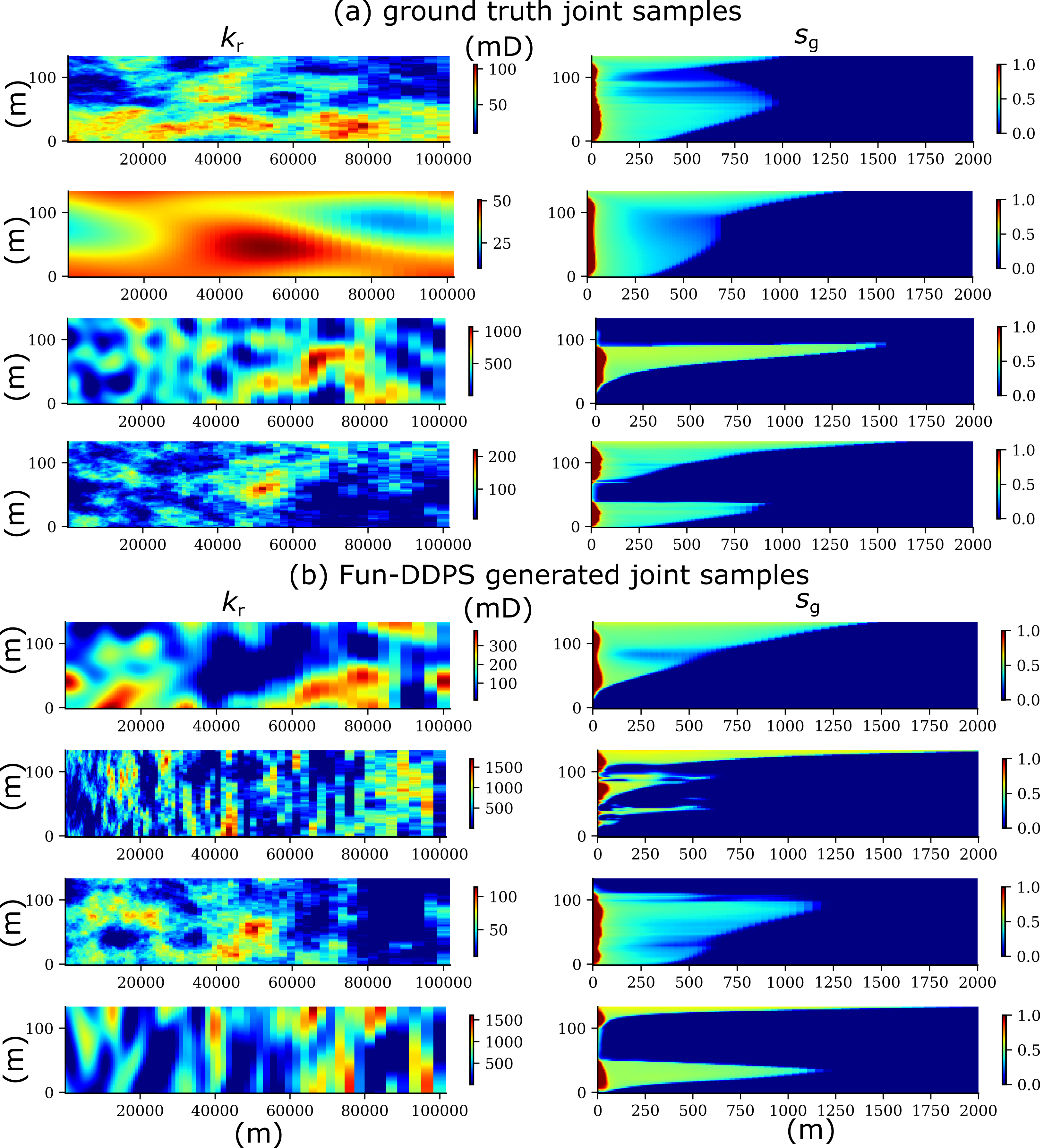}  
    \caption{Randomly selected unconditional samples. (a) Ground Truth: Reference pairs from the test set. (b) Fun-DDPS Generated: Geomodels $\boldsymbol{m}$ generated by the diffusion prior and saturation fields $\boldsymbol{s}$ predicted by the LocalNO surrogate. Note that only the $s_g$ field within $r=2000$ m is shown.}
    \label{fig:unconditional_samples}
\end{figure}

We first assess the visual fidelity of the generated fields (Fig.~\ref{fig:unconditional_samples}).
The diffusion-generated permeability fields $\boldsymbol{m}$ exhibit geologically realistic features, faithfully reproducing the continuous high-permeability channels and spatial heterogeneity characteristic of the training distribution.
Notably, the corresponding saturation plumes $\boldsymbol{s}$---predicted by the surrogate---demonstrate physically consistent migration patterns.
The surrogate successfully captures sharp frontal boundaries and complex fluid-rock interactions driven by the underlying geological heterogeneity, with no observable artifacts.

\subsection{Geological Spatial Statistics: Variogram Analysis}

To quantify the spatial structure of the generated geomodels, we employ variogram analysis.
The semivariogram $\gamma(\boldsymbol{h})$ measures the spatial dissimilarity of the permeability field as a function of lag distance $\boldsymbol{h}$:
\begin{equation}
    \gamma(\boldsymbol{h}) = \frac{1}{2} \text{Var} \left[ \boldsymbol{m}(\mathbf{u}) - \boldsymbol{m}(\mathbf{u} + \boldsymbol{h}) \right]
    \label{equ:two_point}
\end{equation}
where $\text{Var}$ denotes the variance across the ensemble.
Fig.~\ref{fig:unconditional_samples_stats}(a-b) compares the experimental variograms of the generated ensemble against the ground truth.
The Fun-DDPS samples show excellent agreement in both vertical and horizontal directions.
Notably, the uncertainty bands (P5--P95) of the generated ensemble overlap consistently with the ground truth, confirming that the diffusion prior $p_\theta(\boldsymbol{m})$ correctly captures the multi-scale spatial variability of the geological distribution.

\subsection{Dynamic Spatial Connectivity: Two-Point Probability}

We evaluate the induced distribution of dynamic states using two-point connectivity functions on the gas saturation plumes.
This metric calculates the probability that two points separated by a vector $\boldsymbol{h}$ both lie within the CO$_2$ plume ($S_g > 0$), serving as a proxy for plume geometry and connectivity.
As shown in Fig.~\ref{fig:unconditional_samples_stats}(c-d), the statistics of the surrogate-predicted saturation fields closely match the ground truth.
This result is significant: it confirms that the LocalNO surrogate $\mathcal{L}_\phi$ preserves the complex functional mapping from geological heterogeneity to fluid flow, producing saturation fields with statistically correct spatial correlations.

\subsection{Ensemble Distribution Assessment}

\begin{figure}[htbp]
    \centering
    \includegraphics[width=\columnwidth]{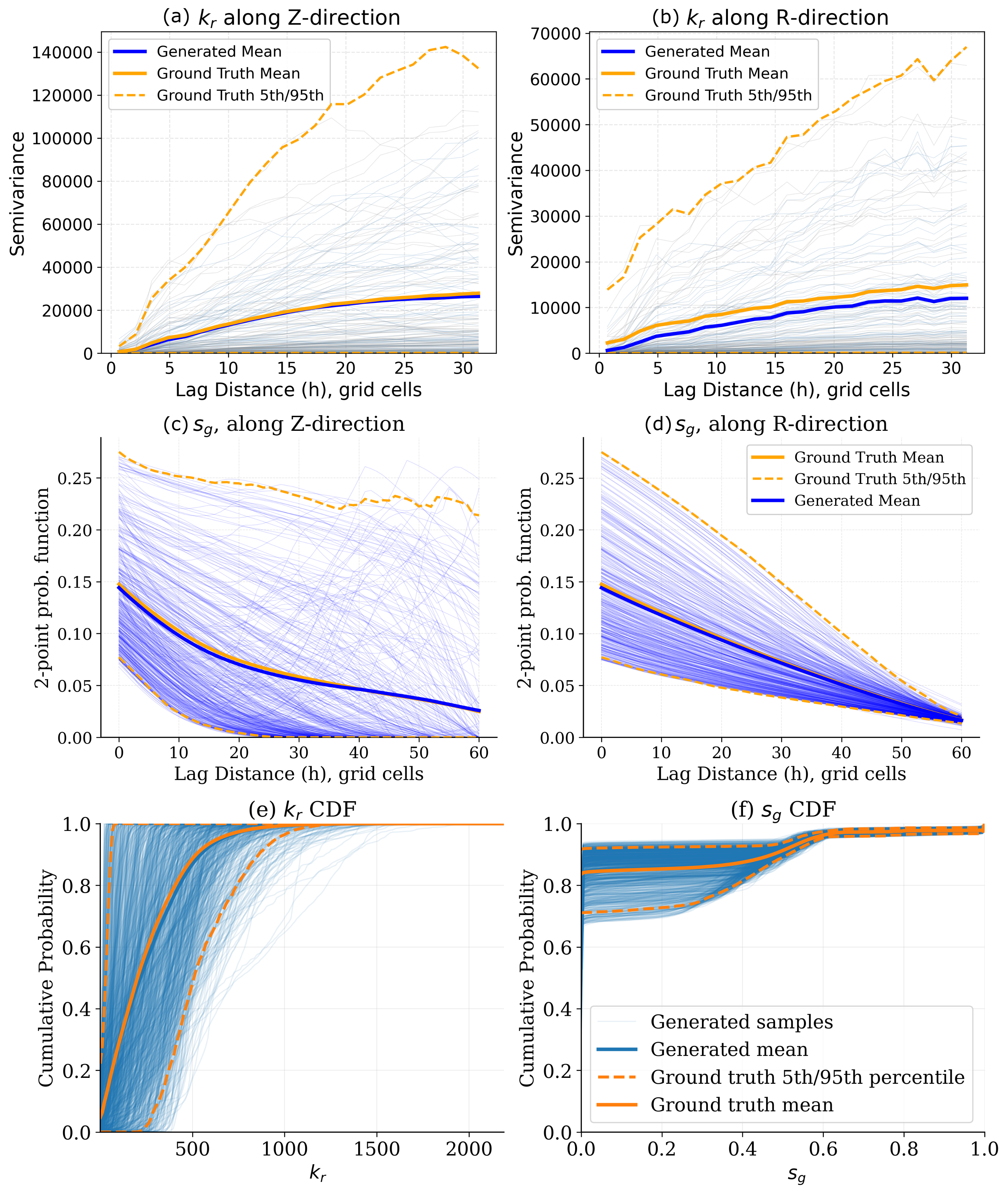}  
    \caption{Quantitative statistical comparison of Fun-DDPS generated samples versus ground truth. (a-b) Experimental variograms for permeability $\boldsymbol{m}$ (validating the diffusion prior). (c-d) Two-point connectivity functions for gas saturation $\boldsymbol{s}$ (validating the surrogate mapping). (e-f) Cumulative distribution functions (CDFs) for both fields.} 
    \label{fig:unconditional_samples_stats}
\end{figure}

Finally, we examine the global statistics via Cumulative Distribution Functions (CDFs) in Fig.~\ref{fig:unconditional_samples_stats}(e-f).
The CDFs for both generated permeability and predicted saturation align closely with the reference distributions.
This confirms that the Fun-DDPS pipeline is unbiased, effectively capturing the full range of geological and dynamical variability present in the training data without mode collapse.

\section{Data Generations and Training Details}
\label{app:pde_training_details}

\subsection{Governing Equations}
The underlying physical system is governed by the conservation of mass and momentum for a two-phase (CO$_2$-water) system in a porous medium. Neglecting chemical reactions and assuming slightly compressible fluids, the mass balance for phase $\alpha$ (where $\alpha \in \{w, g\}$ for water and gas) is given by:
\begin{equation}
    \frac{\partial}{\partial t} (\phi \rho_\alpha S_\alpha) + \nabla \cdot (\rho_\alpha \mathbf{u}_\alpha) = q_\alpha
\end{equation}
where $\phi$ is porosity, $\rho_\alpha$ is density, $S_\alpha$ is saturation, and $\mathbf{u}_\alpha$ is the Darcy velocity vector defined as:
\begin{equation}
    \mathbf{u}_\alpha = - \frac{k_{ri} \mathbf{K}}{\mu_\alpha} (\nabla P_\alpha - \rho_\alpha \mathbf{g})
\end{equation}
The system is closed with the saturation constraint $S_w + S_g = 1$ and capillary pressure relationships $P_c = P_g - P_w$.

\subsection{Simulation Setup}
The domain mimics an infinite-acting aquifer with a radius of 100 km and a thickness of 135 m. We enforce no-flow boundaries at the top and bottom caprock. CO$_2$ is injected at a constant rate of 0.36 Mt/year for 30 years.

\subsection{Data Generation Details}
\label{app:data_gen}

The permeability fields $\boldsymbol{m}$ are generated using Sequential Gaussian Simulation (SGSIM). The geostatistical hyperparameters are sampled from uniform distributions to ensure a wide variety of geological scenarios. The specific ranges are provided in Table~\ref{tab:full_prior_parameters}.

\begin{table}[h]
\centering
\small
\begin{tabular}{lll}
\toprule
Variable & Distribution & Unit \\
\midrule
Radial correlation ($x$) & $\mathcal{U}[359, 35900]$ & m \\
Vertical correlation ($z$) & $\mathcal{U}[14, 56]$ & m \\
Permeability Mean ($\mu_k$) & $\mathcal{U}[10, 500]$ & mD \\
Permeability Std ($\sigma_k$) & $\mathcal{U}[1, 500]$ & mD \\
\bottomrule
\end{tabular}
\caption{Prior distributions for geostatistical parameters.}
\label{tab:full_prior_parameters}
\end{table}

\subsection{Training and Architecture Details}
\label{app:training_details}
We train all diffusion models following the Elucidating Diffusion Models (EDM) framework~\cite{karras2022elucidating}.
The noise level $\sigma$ is sampled during training from a log-uniform distribution over the range $[\sigma_{\min}, \sigma_{\max}] = [0.002, 80]$, ensuring the model learns to denoise effectively across all scales.
Optimization is performed using AdamW~\cite{loshchilov2017decoupled} with a batch size of 64 for 200 epochs.
All experiments were conducted on a single NVIDIA A100 GPU (80GB VRAM).

\subsubsection{Diffusion Operator}
The diffusion model backbone is a U-shaped Neural Operator (U-NO)~\citep{rahman2022u}.
It features a 4-level multi-scale hierarchy (resolutions $64 \times 200 \to 32 \times 100 \to 16 \times 50 \to 8 \times 25$) with a base channel dimension of $d_h=64$ and multipliers $[1, 2, 4, 4]$.
Each resolution level contains residual blocks equipped with spectral convolutions using ComplexTucker factorization (rank $0.1$) for parameter efficiency.
Fourier modes are truncated at 50\% of the spatial resolution at each specific level.
A self-attention mechanism is applied at the $8 \times 25$ bottleneck to capture global dependencies.
Time conditioning $\sigma$ is injected via sinusoidal positional embeddings, which are mapped through an MLP to adaptive scale and shift parameters (AdaGN) within each residual block.
To ensure a fair comparison, both the Fun-DPS joint model (2-channel input $[\boldsymbol{m}, \boldsymbol{s}]$) and the Fun-DDPS geomodel prior (1-channel input $\boldsymbol{m}$) share identical architectural hyperparameters and model size.

\subsubsection{Neural Surrogate}
We employ a Local Neural Operator~\citep{liu2024neural} that combines global Fourier layers with local Discrete Continuous (DISCO) convolutions.
The architecture uses 6 spectral layers with modes truncated at 16 in each spatial direction.
The lifting layer projects inputs to a hidden dimension of $d_h=48$.
DISCO kernels of size $4 \times 4$ are utilized to capture local spatial features while maintaining resolution invariance.

\subsection{Evaluation Metric}
\label{app:eval_metric}

To quantify performance, we compute the relative $L_2$ error between the ground truth solution $\boldsymbol{x}_{\text{gt}}$ and the predicted solution $\boldsymbol{x}_{\text{pred}}$:
\begin{equation}
    \mathcal{E}_{\text{rel}} = \frac{\| \boldsymbol{x}_{\text{gt}} - \boldsymbol{x}_{\text{pred}} \|_2}{\| \boldsymbol{x}_{\text{gt}} \|_2}
    \label{equ:app_rel_l2}
\end{equation}